\documentclass{article}
\usepackage[dvipsnames,table,xcdraw]{xcolor}
\usepackage{amssymb}
\usepackage{booktabs}
\usepackage{tabularx}

\usepackage[final]{corl_2025} 

\usepackage{siunitx}
\usepackage{gensymb}
\usepackage{amssymb}
\usepackage{tcolorbox}
\usepackage{times}
\usepackage{float}
\usepackage{amsfonts}
\usepackage{amsmath}
\usepackage{listings}
\usepackage{mdframed}
\usepackage{soul}
\usepackage{inconsolata}
\usepackage{fancyvrb}
\usepackage{caption}
\usepackage{multirow}
\usepackage{graphicx}
\usepackage{subcaption}
\usepackage[normalem]{ulem}
\usepackage{changepage}
\usepackage{fvextra}
\usepackage{enumitem}
\usepackage{listings}

\lstdefinestyle{LLMQuery}{
  basicstyle=\ttfamily,
  breaklines=true,
  frame=single,
  backgroundcolor=\color{gray!10},
  xleftmargin=0pt,
  columns=fullflexible,
  breakindent=0pt,
  rulecolor=\color{black},
  moredelim=**[is][\textit{}]{\%\%}{\%\%},
  moredelim=**[is][\color{red}\textbf{}\bfseries]{\|\|}{\|\|}
}

\newenvironment{tightlist}%
{\begin{list}{$\bullet$}{%
    \setlength{\topsep}{0in}
    \setlength{\partopsep}{0in}
    \setlength{\itemsep}{0in}
    \setlength{\parsep}{0in}
    \setlength{\leftmargin}{1.5em}
    \setlength{\rightmargin}{0in}
}
}%
{\end{list}
}

\title{CLAMP: Crowdsourcing a LArge-scale in-the-wild haptic dataset with an open-source device for Multimodal robot Perception}

\author{
  \textbf{Pranav N. Thakkar\textsuperscript{\dag}$^1$}, \textbf{Shubhangi Sinha\textsuperscript{\dag}$^1$}, \textbf{Karan Baijal$^1$}, \textbf{Yuhan (Anjelica) Bian$^1$}, \\
  \textbf{Leah Lackey$^1$}, \textbf{Ben Dodson$^1$}, \textbf{Heisen Kong$^1$}, \textbf{Jueun Kwon$^1$}, \textbf{Amber Li$^1$}, \textbf{Yifei Hu$^1$}, \\\textbf{Alexios Rekoutis$^2$}, 
  \textbf{Tom Silver$^1$}, \textbf{Tapomayukh Bhattacharjee$^1$} \\[6pt]
  \textsuperscript{\dag}Equal Contribution \\[6pt]
  $^1$Cornell University, $^2$Horace Mann School
}

\begin{document}
\maketitle

\vspace*{-2.5em}

\begin{center}
    \captionsetup{type=figure}
    \includegraphics[width=1\textwidth]{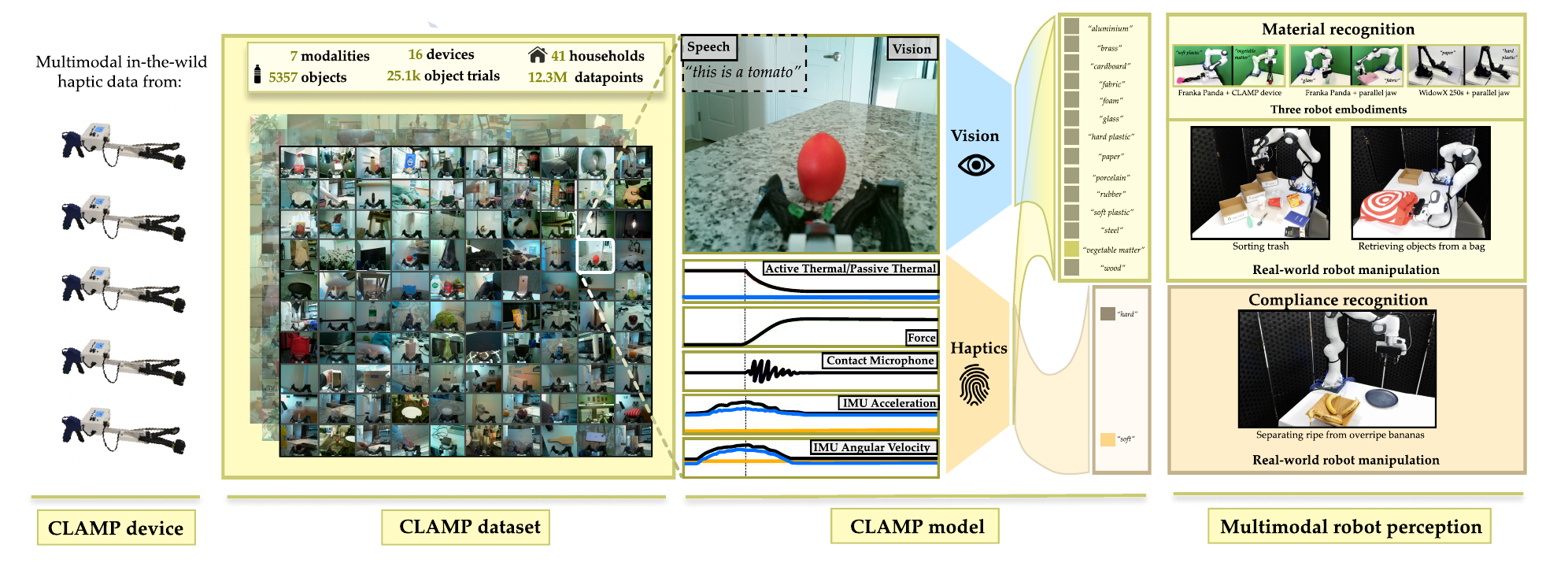}
    \captionof{figure} {We present the CLAMP device for collecting multimodal, in-the-wild haptic data. Using data from 16 devices, we build the CLAMP dataset and train models for material and compliance recognition. Our models generalize to three different robot embodiments and enable real-world robot manipulation.}
    \label{fig:teaser}
\end{center}

\begin{abstract} 
Robust robot manipulation in unstructured environments often requires understanding object properties that extend beyond geometry, such as material or compliance—properties that can be challenging to infer using vision alone. Multimodal haptic sensing provides a promising avenue for inferring such properties, yet progress has been constrained by the lack of large, diverse, and realistic haptic datasets. In this work, we introduce the CLAMP device, a low-cost (\textless \$200) sensorized reacher-grabber designed to collect large-scale, multimodal, in-the-wild haptic data from non-expert users in everyday settings. We aggregate data from 16 CLAMP devices deployed to 41 users to create the CLAMP dataset, the largest open-source multimodal haptic dataset to date, comprising 12.3 million data points across 5357 household objects. Using this dataset, we train a haptic encoder that can infer object material and compliance from multimodal haptic data. We leverage this encoder to create the CLAMP model, a visuo-haptic perception model for material recognition that generalizes to novel objects and three robot embodiments with minimal finetuning. We also demonstrate the effectiveness of our model in three real-world robot manipulation tasks: sorting recyclable waste from trash, retrieving objects from a cluttered bag, and distinguishing overripe from ripe bananas. Our results show that large-scale, multimodal, in-the-wild haptic data collection can unlock new capabilities for generalizable robot manipulation. Device assembly guide, dataset, and code can be found here: \url{https://emprise.cs.cornell.edu/clamp/}


\vspace{-0.5em}

\end{abstract}

\keywords{Multimodal haptic perception,  Data acquisition device, Datasets, Robot manipulation}

\section{Introduction}

Robots benefit from knowledge of object properties beyond geometry, such as material and compliance, to manipulate objects in real-world scenarios such as sorting waste or grasping
deformable objects \cite{arriola2020modeling, bednarek2019robotic, wei2022unknown_material}. However, it is challenging to reliably determine object material or compliance
using vision alone \cite{drehwald2023one, kroemer2019review}. Haptic sensing {helps in} recognizing these object properties, both by itself \cite{chu2013using, chu2015robotic, murali2018learning, dang2012icra} and coupled with vision \cite{gao2016deep, le2021friction}, but it remains less utilized in robotics than vision or language. {This is because state-of-the-art haptic datasets} \cite{yang2022touch, cheng2024touch100k, kerr2022self, bhattacharjee2018multimodal} lack one or more of the following attributes: 

\begin{tightlist}

    \item[1.] \textbf{Multimodality}: Most state-of-the-art haptic datasets capture only contact force \cite{yang2022touch, fu2024touch, calandra2017feeling}. However, haptic sensing is inherently multimodal, including temperature \cite{bhattacharjee2015material, bhattacharjee2016data}, force \cite{bhattacharjee2012haptic, drimus2011classification}, vibration \cite{culbertson2014penn, sinapov2011vibrotactile}, and proprioception \cite{sinapov2014relational}. Prior work has shown that using more haptic modalities improve recognition of object properties such as instance \cite{erickson2017semi} and compliance \cite{bhattacharjee2018multimodal}.

    \item[2.] \textbf{Scalability}: Haptic data requires physical interaction, making {data} collection resource-intensive. Collecting haptic data from multiple users can enable large-scale datasets, but requires standardized sensing hardware that is easy to build and use.
    
    \item[3.] \textbf{Diversity}: Most haptic datasets are collected in controlled environments with a curated list of objects \cite{culbertson2014penn, kerr2022self}. These datasets do not capture real-world haptic interactions during manipulation. Moreover, these datasets do not capture interactions with heterogeneous household objects, i.e., objects made of multiple materials {(e.g. {mobile phones, shoes, books with covers})}. 
    
    \item[4.] \textbf{Data from Grasping Interaction}:
    Haptic sensing is action-conditioned, i.e., it depends on the actions an agent takes while interacting with an object. Most datasets focus on non-prehensile interaction with one virtual finger \cite{chu2015robotic, yu2024octopi, bhattacharjee2018multimodal}. As a result, their data differ from data generated during real-world manipulation involving grasping. 

\end{tightlist}

In this work, we \textbf{c}rowdsource a \textbf{la}rge, multimodal, in-the-wild
haptic dataset for \textbf{m}ultimodal robot \textbf{p}erception. Our key insight is that \emph{{large-scale, multimodal, in-the-wild haptic data} can help to create perception models that enable {real-world} robot manipulation for different embodiments.} Our contributions are as follows:
\begin{tightlist}
    \item[1.] The \textbf{CLAMP device}, a low-cost, sensorized reacher-grabber that captures five haptic modalities: active thermal, passive thermal, force, vibration, and proprioceptive sensing. The device can store data, record image and speech annotations, and is designed for long-term deployment with non-expert users: it is lightweight (0.59 kg), portable, and contains a display with a GUI that allows users to collect data easily with clear instructions.
    \item[2.] The \textbf{CLAMP dataset}, the largest open-source multimodal haptic dataset in the robotics literature — containing 12.3 million data samples from 5357 in-situ household objects gathered by 41 non-expert users with a total of 16 CLAMP devices. The CLAMP dataset contains data from a wide range of household objects with diversity in grasp orientation, speed, and force.
    \item[3.] The \textbf{CLAMP model}, a visuo-haptic model for material classification that generalizes to novel objects and data from multiple robots and gripper types with minimal finetuning. Additionally, the haptic encoder {in} the CLAMP model can be transferred to classify objects based on compliance with no finetuning.
    \item[4.] \textbf{Real robot experiments} that demonstrate a finetuned CLAMP model with a 7-DoF Franka Panda in three real-world robot manipulation tasks: {\underline{\smash{sorting waste into trash and recycling}}}, where the robot reasons about object material {in cluttered scenes}, \underline{\smash{retrieving objects from a}} \underline{\smash{cluttered bag}}, where {it} reasons about object material under occlusion and visual ambiguity, and \underline{\smash{separating ripe from overripe bananas}}, where {it} reasons about object compliance among objects of the same instance.  
\end{tightlist}

\begin{table*}[t]
\centering
\resizebox{\textwidth}{!}{\begin{tabular}{l c c c c c c}
\hline
\textbf{Dataset} & \textbf{Object inst.} & \textbf{Touches} & \textbf{Data Samples} & \textbf{Source} & \textbf{Modalities} \\  
\hline
{Penn Haptic Texture Toolkit \cite{culbertson2014penn}} & 100 & 200 & * & Human (Author) & 3 \\ 
{The Feeling of Success \cite{calandra2017feeling}} & 106 & 9.3k & * & Robot & 2 \\ 
TVL \cite{fu2024touch} & * & * & 44k & Human (Author) & 3 \\
Open Access Haptic Database \cite{bhattacharjee2018multimodal} &  47 & 1340 & * & Human (Author) & 5 \\
Touch100k \cite{cheng2024touch100k} & * & * & 100k & Human (Derived) & 3 \\
Penn Haptic Adjective Toolkit \cite{chu2015robotic} & 60 & 600 & * & Robot & 5 \\
SSVTP \cite{kerr2022self} & * & * & 4500 & Robot & 2 \\
{Proton \cite{burka2018instrumentation}} & 357 & 1.1k & * & Human (Author) & 3 \\ 
Touch and Go \cite{yang2022touch} & 3971 & 13.9k & * & Human (Author) & 2 \\ \hline
\textbf{CLAMP Dataset (Ours)} & \textbf{5357} & \textbf{25.1k} & \textbf{12.3M} & Human (Crowdsourced) & \textbf{7} \\ \hline
\end{tabular}}
\caption{Comparison of the CLAMP dataset with various existing datasets and their characteristics. The CLAMP dataset is the largest open-source haptic dataset to date in terms of data samples and modalities (5 haptic + vision and language) (* indicates unreported figure in source).}
\label{tab:dataset_comparison}
\end{table*}
\section{Related Work}

\textbf{Data acquisition in haptics.} Prior {works} in object instance recognition and material recognition {have} collected haptic data from robots~\cite{chu2015robotic, kerr2022self, calandra2017feeling, calandra2018more, romano2014methods} or human experts~\cite{yang2022touch, kerr2022self, yu2024octopi}, however this approach is constrained by the availability of robots and human experts. While haptic sensing rigs~\cite{proton_2016, wade2015handhelddevicesituacquisition, fu2024touch} provide a pathway to scalable data collection, existing rigs are not optimized for size~\cite{proton_2016}, user-friendliness~\cite{wade2015handhelddevicesituacquisition}, or cost~\cite{fu2024touch}, and also lack streamlined workflows for data collection. In contrast, the CLAMP device is designed {for non-experts to use in their homes.}

\textbf{Haptic datasets.} State-of-the-art haptic datasets~\cite{yang2022touch, cheng2024touch100k, kerr2022self, bhattacharjee2018multimodal, chu2015robotic, kerzel2019neuro,bonner2021dataset, toprak2018evaluating,erickson2017semi, calandra2017feeling, calandra2018more} 
contain fewer data samples than those in vision~\cite{wu2019tencent}, language~\cite{squad} or robot actions~\cite{open_x_embodiment_rt_x_2023}. Previously, Yang et al.~\cite{yang2022touch} created the largest haptic dataset by number of objects, consisting of 3971 objects, while Cheng et al.~\cite{cheng2024touch100k} aggregated the largest dataset by data points with 100k object touches. In comparison, the CLAMP dataset contains 5357 household objects and 12.3 million data samples, representing a {significant} increase in data scale.

Most datasets are collected using a {curated} set of materials~\cite{bhattacharjee2015material, strese2014hapticdatabase, bhattacharjee2016data} or objects~\cite{bhattacharjee2018multimodal, bonner2021dataset, calandra2017feeling}, typically by robots or human experts who ensure uniform interaction and  consistent contact duration~\cite{bhattacharjee2018multimodal, yang2022touch, bonner2021dataset, calandra2017feeling, li2019connecting, chu2015robotic}. {In comparison, our data collection allows users to collect data from any graspable household object}, enabling diverse, in-situ haptic data.
Finally, prior datasets capture some haptic modalities among force~\cite{bhattacharjee2012haptic, bhattacharjee2013rapid, yang2022touch, yu2024octopi, fu2024touch}, vibration~\cite{sinapov2011vibrotactile}, thermal~\cite{bhattacharjee2015material,bai2022analyzing} sensing, {or} a combination of the above~\cite{bhattacharjee2018multimodal, chu2015robotic, kerzel2019neuro, bonner2021dataset, toprak2018evaluating}. In comparison, our dataset captures a total of 7 modalities, including 5 haptic modalities along with vision and structured language.

\textbf{Object property recognition.} {Real-world robot manipulation can benefit from recognizing object properties in the wild \cite{yang2022touch}}. Recognizing object material from vision alone is challenging without more information \cite{bell2015material, drehwald2023one}. Recent works that ground vision–language models in object properties \cite{gao2024physically, agrawal2023physical} also highlight multimodal perception as the next step. While many works have used haptic recognition models to identify various object properties \cite{gao2016deep, chu2015robotic, drimus2011classification}, these models are trained on data from controlled settings. {Yang et al. \cite{yang2022touch} train a material classification model on unimodal, in-the-wild haptic data, specifically high-resolution force data, collected from non-prehensile actions.} In comparison, we train material and compliance classification models on {multimodal, in-the-wild haptic data} collected from grasping actions, enabling recognition for real-world robot manipulation.

{Haptic data is \emph{action-conditioned}, which makes transferring haptic {recognition} models across robots challenging. As a result, most work on object property recognition from visuo-haptic sensing \cite{gao2016deep, erickson2017semi} pretrain or finetune their models on data from only one robot. Yu et al. \cite{yu2024octopi} show that their tactile model, pretrained on {human-collected haptic data}, transfers to classification tasks on robot data, but does not show generalization to multiple robots. Tatiya et al. \cite{tatiya2020haptic} learn a common latent space to transfer haptic knowledge between robots but validate their methods on simulated robots only. In comparison, we show that our visuo-haptic model, pretrained on human-collected data, transfers to three robot embodiments with varying gripper embodiments. }
\section{The CLAMP Device}



\textbf{Sensing.} The CLAMP device features a sensorized reacher-grabber that captures a total of five modalities upon contact with an object. The left sensor bed includes a passive unheated thermal sensor that records the ambient temperature of the object, and an active thermal sensor heated to 55 °C, {whose response over time during contact reflects the object’s heat capacity}. The suction cup on the right end-effector {houses} a contact microphone that measures vibrations, a force-sensing resistor that measures normal force, and a {six-axis IMU}. These sensors are embedded in 3D printed sensor beds as shown in Figure \ref{fig:device-overview}. The device also includes a second IMU, mounted on the device body frame, which helps to isolate the motion of the end-effector. 

\textbf{Compute and storage.}
Our device integrates all sensors with a {Raspberry Pi Pico microcontroller (referred to as Pico) and a Raspberry Pi Zero 2W (referred to as SBC, or single-board computer)}. The Pico is mounted on a custom printed circuit board (PCB) and samples at a rate of 50 Hz for all sensors except the contact microphone, which is sampled at 100 Hz. Data is transferred to the SBC via a Universal Asynchronous Receiver–Transmitter (UART) interface. All data is stored on-board and retrieved when the device is returned after deployment.

\textbf{User interface.}
The SBC is connected to an Adafruit 2.2" PiTFT Display. The SBC takes input from four buttons on the PiTFT and powers the display using PyGame. On booting, the computer loads a graphical user interface (GUI) that allows the user to interact with the device and guides them through the data collection process. 
{Two LEDs, one for each suction cup, provide users with real-time feedback about contact between object surface and each cup, based on sensor readings from the active thermal sensor (right cup) and the force sensor (left cup).}

\textbf{Design.} The device's long handle allows users with limited reach to grasp objects in-situ. We mount the lighter Pico frame along the device length and the heavier SBC enclosure close to the handle, to keep the device ergonomic. As a result, users noted that they held the device in one hand as they repositioned objects with their free hand.

\textbf{Design principles.} We designed the device to be:
\begin{tightlist}
    \item ``\underline{\smash{Easy to Build}}": The CLAMP device can be assembled from scratch within 5 hours, {which is on par with other open-source robotics hardware \cite{zorin2025ruka, shaw2023leap}}. The CLAMP device uses readily available sensors sold with datasheets. We also release all CAD files and assembly instructions.
    \item ``\underline{\smash{Easy to Carry}}":
    The CLAMP device is lightweight (0.59 kg), allowing users to take the device home. We achieved this by selecting the lightweight Pi Zero 2W as our onboard computer and using compact end-effectors that keep the center of gravity of the device close to the handle. 
    \item ``\underline{\smash{Easy to Use}}": Users grasp objects with the reacher-grabber, receive contact feedback via LEDs, and follow on-screen prompts through the GUI to collect data and annotations. Buttons on the PiTFT allow users to track progress and power down the device {when not in use}.
    \item ``\underline{\smash{Easy to Scale}}": One CLAMP device costs $<\$200$ to build, {making it more affordable than existing haptic sensors \cite{digit, digit360, gelsight} and sensing suites \cite{clarke2025xcapture}}. The sensorized suction cups can be mounted on robot end-effectors, allowing roboticists to collect haptic data with their own robots.
\end{tightlist}


\vspace{-1.0em}
\section{The CLAMP Dataset}
\label{sec:dataset}
\vspace{-0.3em}

The \textbf{CLAMP dataset} is the largest open-source multimodal haptic dataset to date. The dataset consists of 5357 object instances, 25.1k object touches, and 12.3M individual data samples. We gathered this data from 16 CLAMP devices {deployed to 41 users}. 

\textbf{Data collection process.} {Each user is given a short tutorial and a manual on how to use the device. The GUI guides users through three steps of data collection for an object: capturing an image of the object, recording a spoken annotation, and collecting haptic data by grasping the object {for up to 10 seconds per trial}, for five trials. The GUI then returns to the start screen. At the end of the deployment, the device is returned for maintenance and data backup.}



\textbf{Data post-processing and featurization.} {We collect data from six haptic sensors during each trial: active and passive thermal sensors, a force sensor, a contact microphone, and two IMUs. After synchronizing all sensor data streams, we apply smoothing filters and extract features {from each sensor, resulting in a total of nine features. Refer to Appendix \ref{appendix:data_processing} for more details.}
We synchronize contacts for each {suction cup} by detecting  contact onset and loss for the left and right cups separately. We elaborate on the rule-based approach used to {detect contact onset and release for each cup}, in Appendix \ref{appendix:synchronizing_contact}. Finally, we pad the features to ensure all features are of the same length.

\textbf{Object property annotations.} We generate annotations for object material and compliance to enable model learning with the CLAMP dataset. In particular, each object is labeled with:
\begin{tightlist}
    \item[1.] One of 16 \textbf{material labels}: \texttt{\{aluminium, brass, cardboard, dry wall, fabric, foam, glass, granite, hard plastic, paper, porcelain, rubber, soft plastic, steel, vegetable matter, wood\}}. {We create this list from 300 material subcategories from the CES Edupack \cite{cesedupack2023} database that appear in our dataset, which we group by similarity towards downstream manipulation. Our material list covers $93\%$ of objects in our dataset.}
    \item[2.] {If the object has different materials on opposing surfaces, i.e., \textbf{ heterogeneous surfaces}: \texttt{\{yes, no\}}. Note the difference between these objects (e.g. mobile phones) and \emph{heterogeneous objects} (multiple materials overall, e.g. furniture).}
    \item[3.] An adjective label describing \textbf{object compliance}: \texttt{\{soft, hard\}}.
\end{tightlist}

To generate these labels, we first transcribe the audio annotation using 
Whisper~\cite{radford2023robust}. We then provide the audio transcription along with the image annotation for the object to GPT-4o \cite{gpt4technicalreport}, along with more than 12 in-context examples. We detail the exact prompts used in Appendix \ref{appendix:data_annotation}. Finally, expert annotators verify the audio transcription and material labels and re-run the annotation pipeline for any corrections made.

\textbf{Dataset statistics.} 
The CLAMP dataset contains diversity in objects and in grasping actions. In the appendix, we quantify diversity across three axes: materials, grasping forces, and grasping speeds. Refer to Appendix \ref{appendix:dataset_statistics} for more details. 

\begin{figure*}[tbp]
\centering
\includegraphics[width=1\linewidth]{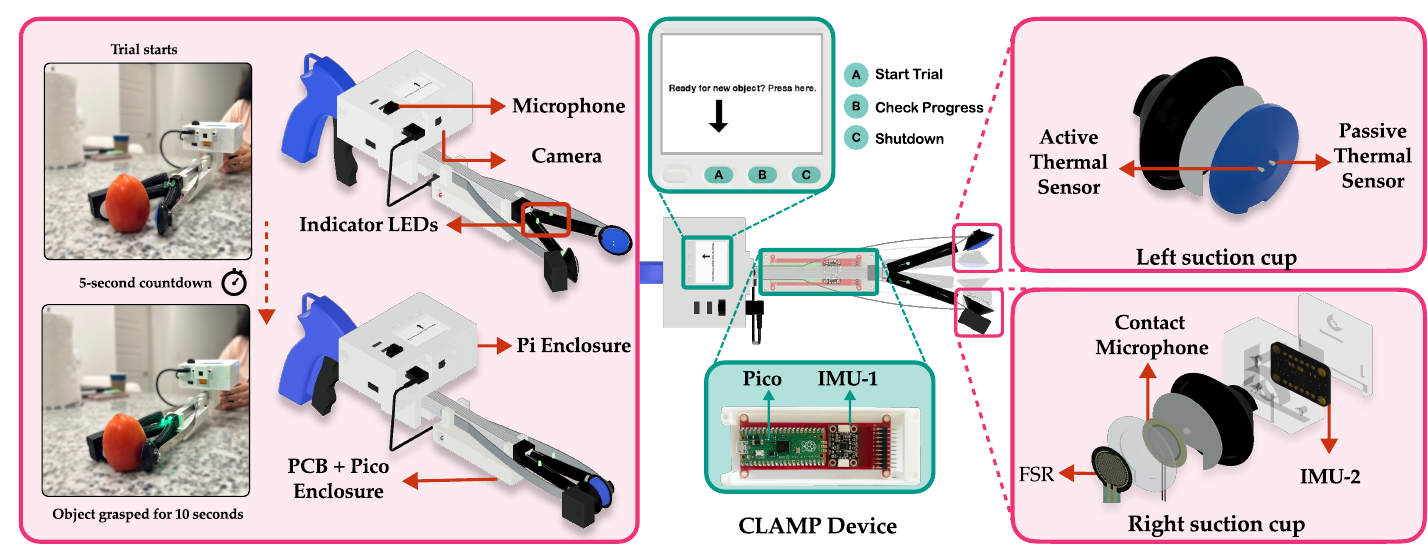}
\caption{\textbf{Device overview: } The CLAMP device features a modified reacher-grabber equipped with
sensors that capture five haptic modalities. Our device enables non-expert users to collect haptic data in their homes.}
\label{fig:device-overview}
\end{figure*}

\section{The CLAMP Model}

\label{sec:clamp-model}{
{The \textbf{CLAMP model} is a {multimodal} perception model} that recognizes object material from visual and multimodal in-the-wild haptic data.
The model fuses a haptic encoder trained on the CLAMP dataset with a pretrained visual encoder.}

\begin{figure*}[ht]
\centering
\includegraphics[width=1\linewidth]{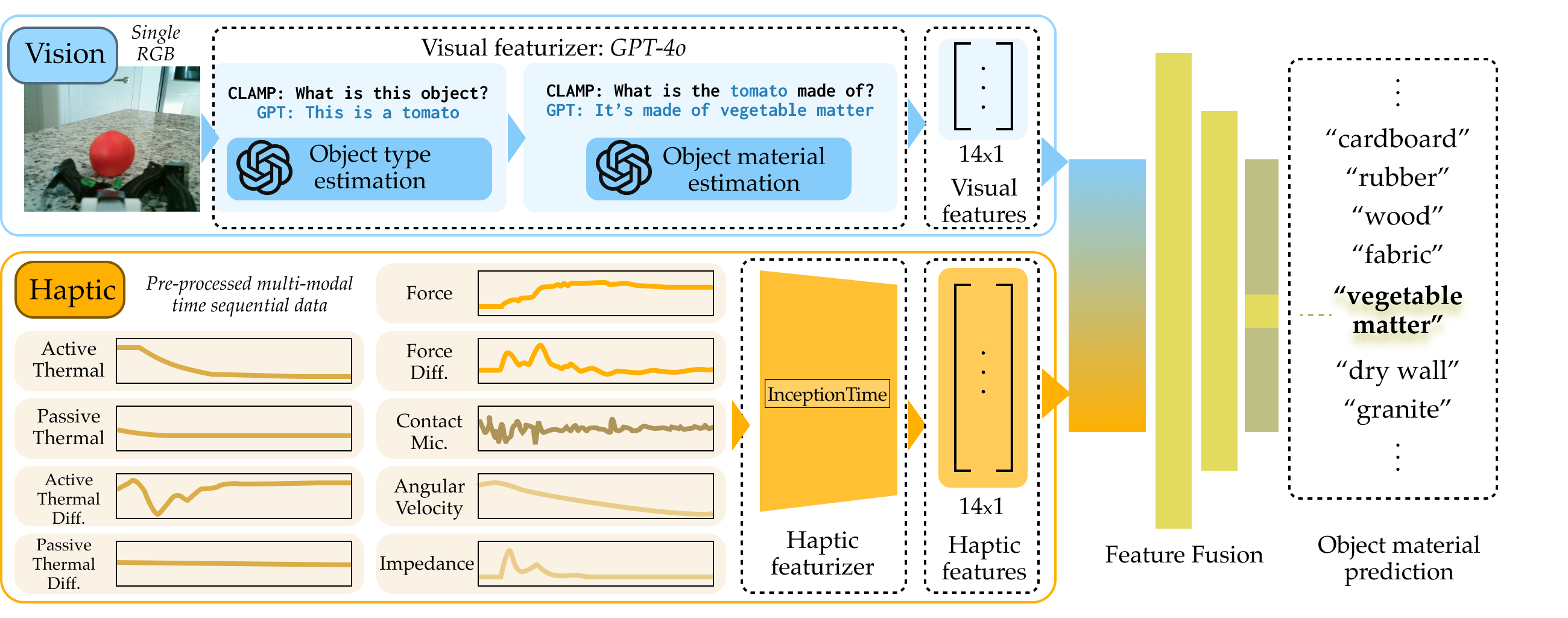}
\caption{\textbf{Model overview: } We propose the CLAMP model, a visuo-haptic model that fuses outputs from a GPT-4o~\cite{gpt4technicalreport} visual encoder and a pretrained InceptionTime-based~\cite{ismail2020inceptiontime} haptic encoder.}
\label{fig:model-overview}
\end{figure*}

\textbf{Haptic Encoder.} {Our haptic encoder is a six-block InceptionTime model ~\cite{ismail2020inceptiontime}, followed by a global average pooling layer, and finally a three‐layer MLP as the classification head. 
We split the CLAMP dataset sample-wise, train the encoder over three random seeds, and use the worst-performing model for comparisons. See Appendix \ref{appendix:model_training} for more details.}

\textbf{Visual Encoder.} 
We use GPT-4o \cite{gpt4technicalreport} as a low-dimensional visual encoder. {Given an input image, audio transcription, and in-context examples, it outputs per-token probabilities, which we filter and normalize into a categorical distribution over material classes.} See Appendix \ref{appendix:features_gpt} for more details.




\textbf{The CLAMP Model: Combining Haptics and Vision.} The CLAMP model uses a 2-layer MLP to fuse the outputs of the haptic and visual encoders.
To train the model, we freeze the pretrained haptic encoder and minimize a loss function: $\mathcal{L} = \mathcal{L}_{WCE} + \lambda_{KL} \cdot \mathcal{L}_{KL} (\mathcal{V} || \mathcal{P})$ , where $\mathcal{L}_{WCE}$ is a weighted cross-entropy loss,  $\mathcal{L}_{KL} (\mathcal{V} || \mathcal{P})$ is a KL-divergence loss term with the vision probabilities $\mathcal{V}$ and the visuo-haptic model probabilities $\mathcal{P}$, {and $\lambda_{KL}$ is the weight for KL loss} (Appendix \ref{appendix:model_training}).



\textbf{Comparison to Pretrained Vision Foundation Models.} We compare the CLAMP model on material classification performance to GPT-4o~\cite{gpt4technicalreport}, a grounded vision-language model (PG-VLM~\cite{gao2024physically}), and a CLIP~\cite{radford2021learning} model finetuned on our dataset. For each model, we consider three inputs:
\underline{\smash{raw images}}, images \underline{\smash{cropped}} to the object instance using Grounding DINO \cite{liu2025grounding}, and images \underline{\smash{segmented}} by Grounded SAM \cite{ren2024grounded}. We compare the CLAMP model with the three vision-only models, across all inputs, in  
Table~\ref{tab:all_results}, {which reports test-set accuracy and normalized Matthews correlation coefficient (nMCC)}. The CLAMP model achieves the best performance, suggesting that the background knowledge of the pretrained models cannot compensate for their lack of haptic input.

\textbf{Haptic Modality Ablation Studies.}
{We compare classification performance of the haptic encoder to ablated encoders trained on all but one haptic modality.} In Table~\ref{tab:all_results}, we see that the encoder trained on all modalities performs best, suggesting the importance of each modality. All five modalities provide complementary cues about object properties. {Force conditioned on proprioception distinguishes soft objects from hard ones} \cite{bhattacharjee2018multimodal}, vibration sensing captures high-frequency contact cues in hard objects, and active/passive thermal sensing separates objects of similar compliance by {thermal} conductivity and temperature \cite{bai2022analyzing}. Finally, we see that the CLAMP model performs better than the haptic encoder alone, confirming that both vision and haptics are necessary for material recognition. 

\textbf{Alternative Haptic Encoders.} {
{We} consider alternative network architectures for the haptic encoder.}
We first compare our InceptionTime encoder to a Random Forest encoder used in previous work on haptic classification~\cite{da2022tactile}, finding InceptionTime to obtain better performance (Table~\ref{tab:all_results}).}

\textbf{Visuo-Haptic Fusion Strategy.} {We explore finetuning CLIP and fusing it with the haptic encoder. While CLIP underperforms in comparison with GPT-4o in Table~\ref{tab:all_results}, it enables learning a visuo-haptic representation with high-dimensional embeddings. We compare the CLAMP model to a setup where a finetuned CLIP encoder is fused with the haptic encoder, using high-dimensional embeddings from both. The CLAMP model outperforms the CLIP-based fusion, suggesting that fusing higher-dimensional features does not necessarily lead to better performance.}

\textbf{Testing on Held-out Objects.} {We train a new CLAMP model on a dataset split into training, validation, and test sets by objects rather than by contact instances. The CLAMP model outperforms the vision-only baseline on the new test set, suggesting that our model can infer object material when tested on novel objects. {The CLAMP model’s 3\% accuracy improvement over the vision-only baseline is statistically significant with $p \:\textless \: 0.05$ (using the Wilcoxon-Signed Rank test)}.

\textbf{Transferring the Haptic Encoder to Compliance Recognition.}
We show that our haptic encoder, which was trained for material classification, can be transferred to recognize object compliance. We train a new classification head on the frozen haptic encoder using object compliance recognition labels (\texttt{soft} and \texttt{hard}) from the CLAMP dataset.
The haptic encoder achieves an accuracy of 0.88 on the test set and an nMCC score of 0.869. This result suggests that our haptic encoder learns representations that are applicable beyond material recognition.

\textbf{Analyzing CLAMP Model Uncertainty.} Uncertainty quantification is important for robot manipulation tasks to reduce failures ~\cite{ren2023robots}.
To assess whether CLAMP model predictions carry uncertainty signals, we mark a prediction as \emph{uncertain} if the corresponding softmax value is less than a threshold $p_1$, or if the difference between the two largest softmax values is less than a threshold $p_2$. {Filtering predictions increases the model test accuracy from 0.73 to 0.82, and the nMCC from 0.85 to 0.90, while retaining $75\%$ of predictions.} {Refer to Appendix \ref{appendix:model_training} for values of $p_1$ and $p_2$.} 

\begin{table*}[t]
  \centering
  \scriptsize
  \setlength{\tabcolsep}{4pt} 
  \begin{tabular}{%
        @{}                                     
        >{\centering\arraybackslash}p{0.35\textwidth}
        | >{\centering\arraybackslash}p{0.30\textwidth}
        | >{\centering\arraybackslash}p{0.29\textwidth}
        @{}                                     
    }
    \begin{tabular}[t]{@{}lcc@{}}
      \toprule
      \multicolumn{3}{c}{\bfseries Haptic Models} \\
      \midrule
      Method           & Acc.    & nMCC     \\
      \midrule
      \textbf{InceptionTime}    & \textbf{0.59} $\pm$ \textbf{0.01} & \textbf{0.76} $\pm$ \textbf{0.006} \\
      Random Forest    & 0.52 $\pm$ 0.001& 0.71 $\pm$ 0.001 \\
      \bottomrule
      \toprule
      \multicolumn{3}{c}{\bfseries Modality-specific Ablation} \\
      \midrule
      Absent Modality           & Acc.    & nMCC     \\
      \midrule
      \textbf{\emph{None}}             & \textbf{0.58}  & \textbf{0.76} \\
      Active Therm.    & 0.49  & 0.7  \\
      Passive Therm.   & 0.46  & 0.68 \\
      Force            & 0.22  & 0.57 \\
      Vibration        & 0.57  & 0.75 \\
      Proprioception   & 0.54  & 0.74 \\
      \bottomrule

    \end{tabular}
    &
    \begin{tabular}[t]{@{}lcc@{}}
      \toprule
      \multicolumn{3}{c}{\bfseries Vision Models} \\
      \midrule
      Method             & Acc.  & nMCC   \\
      \midrule
      \textbf{GPT-4o} (raw)       & \textbf{0.65}  & \textbf{0.80}   \\
      GPT-4o (segmented) & 0.59  & 0.77   \\
      GPT-4o (cropped)   & 0.64  & 0.80   \\
      \midrule
      \textbf{CLIP (raw)}         & \textbf{0.58}  & \textbf{0.75}   \\
      CLIP (segmented)   & 0.56  & 0.74   \\
      CLIP (cropped)     & 0.57  & 0.74   \\
      \midrule
      PG-VLM (raw)       & 0.39  & 0.67   \\
      PG-VLM (segmented) & 0.44  & 0.69   \\
      \textbf{PG-VLM (cropped)}   & \textbf{0.58}  & \textbf{0.76}   \\
      \bottomrule
    \end{tabular}
    &
    \begin{tabular}[t]{@{}lcc@{}}
      \toprule
      \multicolumn{3}{c}{\bfseries Visuo-haptic models} \\
      \midrule
      Method                  & Acc.  & nMCC   \\
      \midrule
      \textbf{CLAMP model}              & \textbf{0.87}  & \textbf{0.93}   \\
      CLIP + Haptic encoder    & 0.79  & 0.89   \\
      \bottomrule
      \toprule
      \multicolumn{3}{c}{\bfseries Evaluation on held-out objects} \\
      \midrule
      Method                  & Acc.  & nMCC   \\
      \midrule
      \textbf{CLAMP model}              & \textbf{0.73}  & \textbf{0.85}   \\
      GPT-only                 & 0.70  & 0.83   \\
      \bottomrule
    \end{tabular}
  \end{tabular}
  \caption{We report the performance of material recognition models on the CLAMP dataset. Note that for ``Haptic Models", metrics are reported as mean $\pm$ stddev. }
  \label{tab:all_results}
\end{table*}

\section{Demonstrating the CLAMP Model on Real Robots}

\begin{figure*}[h]
\centering
\includegraphics[width=1\linewidth]{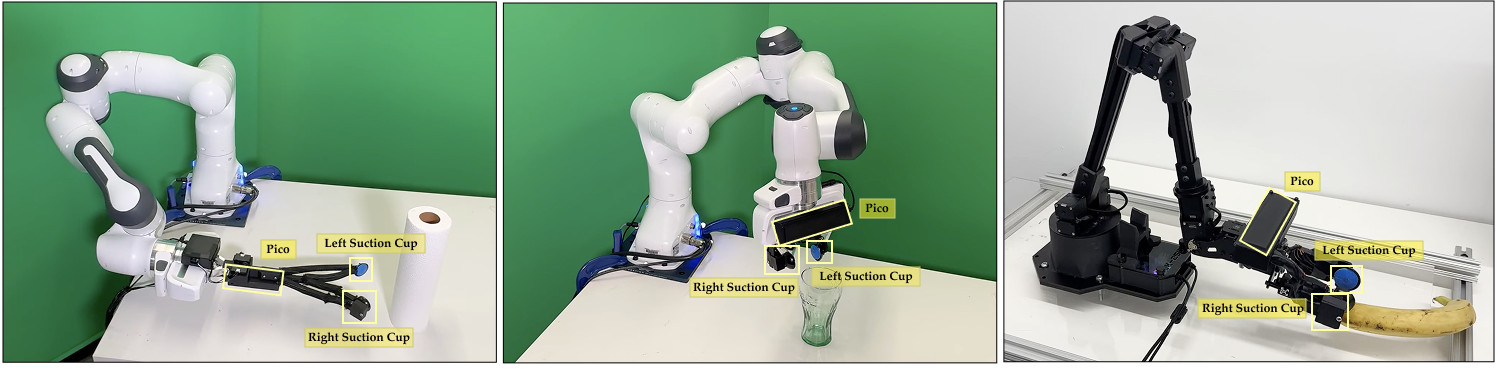}
\caption{We mount haptic sensors from the CLAMP device on three robot embodiments and collect haptic data. \textbf{Left.} Franka with CLAMP device. \textbf{Middle.} Franka with Parallel Jaw. \textbf{Right.} WidowX with Parallel Jaw.} 
\label{fig:robot-embodiments}
\end{figure*}

In this section, we demonstrate that the CLAMP model enables multimodal robot perception for multiple embodiments and thus improves robot manipulation capabilities in real-world scenarios.

\textbf{Finetuning the CLAMP model on robot data.} We collect haptic data from three different robot embodiments, and show that a pretrained CLAMP model requires little finetuning data to exceed the performance of vision-only methods towards material recognition. We consider three embodiments:

\begin{tightlist}
    \item[1.] \underline{Franka with CLAMP Device}: We mount a shortened CLAMP device on the end effector of the Franka Panda. The device is actuated by tensioning its central steel strips, which are connected to a pulley system driven by a pair of interlocked Dynamixel XC330-M288-T motors. A 3D printed housing contains the motors and the Pico. 
    \item[2.] \underline{Franka with Parallel Jaw}: We directly attach the suction cups of the CLAMP device to the fingers of the Franka arm using 3D printed mounts. An enclosure on the Franka Hand houses the Pico.
    \item[3.] \underline{WidowX with Parallel Jaw}:  We similarly directly attach the suction cups of the CLAMP device to the fingers of a WidowX 250s arm using 3D printed mounts. An enclosure houses the Pico.
\end{tightlist}

{The narrow grasping width for the two parallel jaw embodiments prevents grasps for several household objects, resulting in a smaller dataset for these embodiments compared to that for the first one.} We post-process and featurize the collected haptic data following the steps described in Section~\ref{sec:dataset}. However, for embodiments with a parallel jaw gripper, we modify the impedance feature calculation to use linear acceleration instead of angular velocity. Refer to Appendix \ref{appendix:robot_finetuning} for details.

We finetune a pretrained CLAMP model with varying amounts (7\%, 15\%, and 30\% data) of embodiment-specific haptic data and offline images collected with a CLAMP device. Table \ref{tab:robot_data} shows that using more finetuning data improves model performance, and that finetuned models for all three embodiments perform better than vision-based material predictions with $15\%$ of finetuning data.

\vspace*{-1mm}

\begin{table*}[h]
  \centering
  \scriptsize
  \setlength{\tabcolsep}{4pt} 
  \begin{tabular}{%
        @{}                                     
        >{\centering\arraybackslash}p{0.28\textwidth}
        | >{\centering\arraybackslash}p{0.27\textwidth}
        | >{\centering\arraybackslash}p{0.27\textwidth}
        @{}                      
    }
    \begin{tabular}[t]{@{}lcc@{}}
      \toprule
      \multicolumn{3}{c}{\bfseries Franka with CLAMP device} \\
      \midrule
      Finetuning Data (\%)           & Acc.    & nMCC     \\
      \midrule
      \emph{Vision-only}  & 0.63  &    0.80  \\
      Zero-shot     & 0.55 & 0.76 \\
      7$\%$         & 0.73 & 0.85 \\
      15$\%$        & 0.77 & 0.88 \\
      30$\%$        & 0.82 & 0.90 \\
      \bottomrule
    \end{tabular}
    &
    \begin{tabular}[t]{@{}lcc@{}}
      \toprule
      \multicolumn{3}{c}{\bfseries Franka with Parallel Jaw} \\
      \midrule
      Finetuning Data (\%)           & Acc.    & nMCC     \\
      \midrule
      \emph{Vision-only}  &  0.77    &    0.87     \\

      7$\%$         & 0.75 & 0.86 \\
      15$\%$        & 0.93 & 0.96 \\
      30$\%$        & 0.95 & 0.97 \\
      \bottomrule
    \end{tabular}
    &
    \begin{tabular}[t]{@{}lcc@{}}
      \toprule
      \multicolumn{3}{c}{\bfseries WidowX with Parallel Jaw} \\
      \midrule
      Finetuning Data (\%)           & Acc.    & nMCC     \\
      \midrule
      \emph{Vision-only} & 0.68    &    0.83  \\
      7$\%$         & 0.71 & 0.83 \\
      15$\%$        & 0.71 & 0.84 \\
      30$\%$        & 0.81 & 0.89 \\
      \bottomrule
    \end{tabular}
  \end{tabular}
  \caption{We report the performance of the CLAMP model finetuned on robot-collected data for three robot embodiments, for varying amounts of finetuning data.}
  \label{tab:robot_data}
\end{table*}


\textbf{Leveraging multimodal perception for real-world manipulation.} We finetune a CLAMP model with 85\% of the finetuning dataset for the ``Franka with Parallel Jaw'' embodiment equipped with a wrist camera, and deploy it in three real-world scenarios.

\begin{figure*}[h]
\centering
\includegraphics[width=1\linewidth]{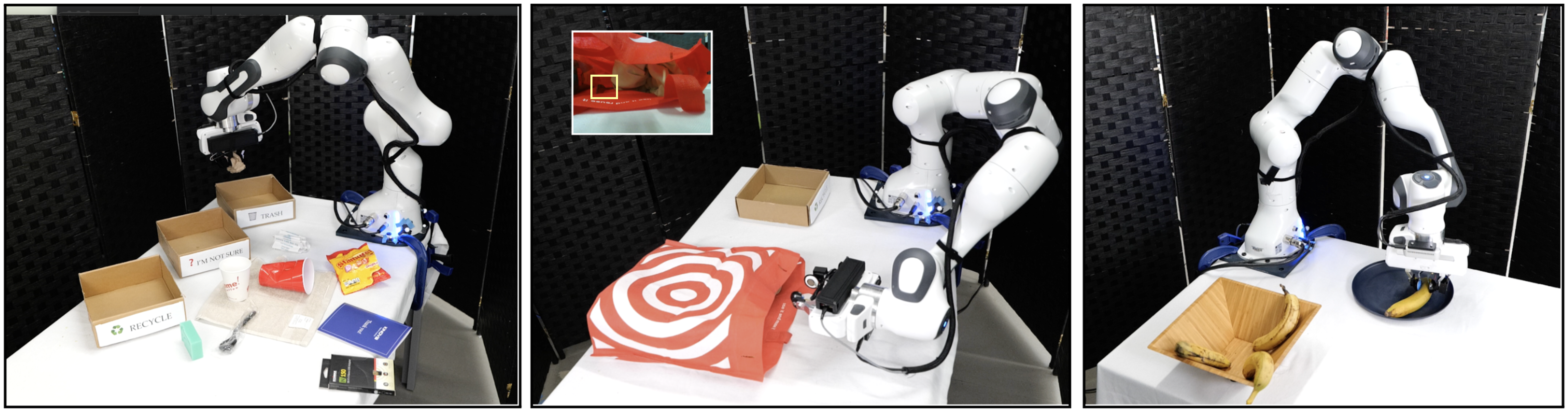}
\caption{We demonstrate the CLAMP model in three real-world robot manipulation scenarios. \textbf{Left.} The robot sorts recyclables from non-recyclables. \textbf{Middle.} The robot retrieves a metallic object (in yellow square) from a bag with multiple objects. \textbf{Right.} The robot distinguishes overripe from ripe bananas.} 
\label{fig:robot-manip}
\end{figure*}

\begin{tightlist}

\item \underline{\smash{Sorting recyclable waste from trash}}: 
In this setting, the robot disposes of 10 objects not seen in the finetuning dataset, into ``Trash'' or ``Recycle'' bins based on material predictions from the CLAMP model (sorting rules detailed in Appendix \ref{appendix:robot_sorting}), {with an ``I'm not sure" box for uncertain predictions}. The robot disposes of objects in the correct bin 9/10 times per trial, while generating unknown predictions 0.66/10 times per trial, over three trials. {This shows that our model can accurately recognize material for unseen objects in cluttered environments.}

\item \underline{\smash{Retrieving objects from a cluttered bag}}:
{In this setting, the robot interacts with objects {inside} a cloth bag, and must retrieve the object only if it is metallic.
The bag contains three unseen objects, one of which is metallic. We also log predictions from the vision-only model for comparison. The finetuned CLAMP model retrieved the metallic object in 6 out of 13 trials, while the vision model does not find a metallic object in any of the 13 trials. We thus show that haptics provides key cues in visually challenging situations. See Appendix \ref{appendix:robot_bag} for more details.}

\item \underline{\smash{Sorting ripe from overripe bananas}}: {In this setting, the robot interacts with four bananas (two of which are overripe), predicts its compliance, and retrieves only ripe bananas. Our haptic encoder transferred to compliance prediction and finetuned on robot data achieves an accuracy of 0.833 and an nMCC of 0.86. This shows that our model can differentiate between visually ambiguous objects of the same instance, based on object compliance. }

\end{tightlist}

\section{Limitations and Future Work}

In this work, we present the CLAMP device, dataset, and model, which collectively represent a significant step towards scaling visuo-haptic perception for robotics.
We conclude with a discussion of limitations in the current framework and opportunities for future work.

\textbf{CLAMP Device Design Constraints. } The CLAMP device was optimized for weight, cost and portability. This lightweight design came with trade-offs. For example, we opted against using a more powerful microprocessor like the Raspberry Pi 4, which would have enabled higher bandwidth data collection such as video streams, high-frequency contact microphone data, and vision-based tactile sensing. In addition, the current suction cup design allows for agile handling, but its low surface area and convex sensor bed can hinder consistent contact with objects with curved surfaces. 

\textbf{CLAMP Dataset Limitations.} 
While the CLAMP dataset is the largest haptic dataset to date, it is still far smaller than counterparts in vision and language. It remains to be seen how much data is required to train large visuo-haptic foundation models that have emergent generalization capabilities analogous to large language and vision-language models. Furthermore, while the CLAMP dataset features diversity along multiple axes, it does not have diversity in terms of sensors used for data collection. Future work involves combining our dataset with data from other sensors, such as vision-based tactile sensors \cite{digit}.

\textbf{CLAMP Model Limitations. } 
The CLAMP model performs better than vision-only and haptic-only perception models, but there is much room for future work on learning better visuo-haptic models with the CLAMP dataset.
A direction to explore in future work is training larger models that operate directly on the raw visuo-haptic inputs.
The structure we imposed by featurizing the haptic data and prompting a VLM for visual encoding was helpful for the current CLAMP model, but this may change as data and compute continue to scale.


\acknowledgments{This work was partly funded by National Science Foundation IIS \#2132846, and CAREER \#2238792. We thank Ruolin Ye, Rishabh Madan, Qi Chen, and other members of the EmPRISE Lab for feedback on the CLAMP device, assistance with figures in the manuscript, and assistance in data collection.}


\bibliography{references}  

\clearpage
\appendix

{\Large{\textbf{Appendix}}}

\section{The CLAMP device}

\subsection{Hardware details}
\label{appendix:hardware}

The details of the sensors and peripherals onboard the CLAMP device are provided below:

\textbf{Active and passive thermal sensing.}
The CLAMP device uses two 10 k\si{\ohm} B57541G1103F NTC thermistors for active and passive thermal sensing.
The active thermal sensor is maintained at a temperature of 55 °C. The change in active thermal sensor readings over time, when the sensor comes in contact with an object, is indicative of the heat capacity of the object.
The passive thermal sensor measures the surface temperature of the object in contact. 

\textbf{Force sensing.}
The CLAMP device uses an Interlink UX 402 force-sensing resistor (FSR) that can measure forces up to 150 N.
We calibrate the force-sensing resistor with an FX29 load cell. The resulting calibration curve is described by an exponential function of the voltage, achieving an R2 = 0.980. While not as accurate as load cell sensors or MEMS force sensors over long-term cyclic use, force-sensing resistors offer significant advantages in terms of resilience and cost.

\textbf{Vibration sensing.}
The CLAMP device uses a 20 mm diameter piezo disc, also known as a contact microphone, and a MAX4466 amplifier with a gain of 25 to measure audio signals resulting from contact.

\textbf{Proprioceptive sensing.}
An important aspect of haptics is that sensing depends on action. The contact forces generated while grasping an object depend on the velocity and the angle at which the object is grasped. To address this, the CLAMP device has two 6-axis MPU6050 IMUs.
The axes of the two IMUs are oriented such that the Y-axis of IMU-2 aligns with the Z-axis of IMU-1, while the x-axis of IMU-2 is transformed at an angle of $-25\degree$ from that of IMU-1, about the Z-axis of IMU-1.

\section{The CLAMP dataset}

\subsection{Data processing}
\label{appendix:data_processing}

The raw haptic signals from the CLAMP device are processed in two stages: first, they are smoothed to produce visually interpretable data; then, features are extracted for model learning. We outline the exact details below:
\begin{enumerate}
    \item We first align all sensor data (from the active and passive thermal sensor, the force sensor, the contact microphone, and the two IMUs) based on their timestamps,  ensuring that readings are synchronized to within 2 ms.
    \item We convert raw voltage readings from the active and passive thermal sensors to °C using the resistor values provided in the datasheet. \footnote{Despite access to a curve fit obtained from calibration, we do not convert raw voltage readings to readings in Newtons for the FSR. We train our model on uncalibrated force values. We verified that a material recognition model trained on calibrated force values performs equally as well as one trained on uncalibrated force values. }
    \item Data from some modalities are filtered to remove noise:
    \begin{itemize}
        \item \underline{\smash{Active and passive thermal sensor}} data is passed through a Butterworth filter.
        \item \underline{IMU data} is passed through a causal moving average filter.
    \end{itemize}
    
\end{enumerate}

This results in haptic data that is visually interpretable.
\begin{enumerate}[start=4]

\item We extract a total of nine features from all sensor data streams. For the thermal sensors, the force sensor, and the contact microphone, we use the smoothed readings directly. For the IMU data, we compute the angular velocity of the end-effector IMU relative to the IMU in the gripper body. This results in five primary features.
We then create four additional features: differences in active thermal, passive thermal, and force readings between consecutive timesteps, and an impedance feature. Note that:


\begin{enumerate}
    \item For angular velocity feature, we compute the gyroscope readings of IMU-2 relative to motion of IMU-1, in the frame of IMU-1. We consider the relative angular velocity of IMU-2 about the Z-axis of IMU-1 as a feature. 
    \item For impedance, we use the following formula:
\begin{align*}
\label{impedance_eqn}
        Z(t) &= \begin{cases}
                    \dfrac{F'(t)}{\omega(t)} \;\;\; &\text{if } \omega(t) \geq \delta \\[5pt]
                    0 \;\;\; &\text{if } \omega(t) < \delta
                    \end{cases}
\end{align*}
where $F'(t)$ denotes the force difference at time $t$, $\omega(t)$ denotes the angular velocity feature, and $\delta$ denotes a threshold for angular velocity, which we fix as $3 ^\circ / s$ for our experiments. We find that this impedance feature reliably indicates contact and shows distinct responses for contact with rigid and compliant objects.
While impedance is typically computed via linear velocity, we use angular velocity because our IMU-based proprioception is more accurate in this dimension.
We fix a lower bound to exclude spurious values of high impedance that we observe at low angular velocities. This often happens when users attempt to change the grasp contact point without releasing their grasp, resulting in a sudden change in force measurement from the FSR.
\end{enumerate}

\item We apply a causal moving average filter on each feature except the contact microphone. The contact microphone readings are debiased and then downsampled to match the length of the other features.

\item Finally, the length of all features is set to 491 timesteps. Shorter features are padded with the last value or 0, depending on the feature. 

\end{enumerate}

\subsection{Synchronizing contact}
\label{appendix:synchronizing_contact}

Contact is defined differently for sensors, depending on the suction cup they are located on. We synchronize contact for both cups by using a rule-based approach for sensors on each cup. 

\begin{itemize}
    \item Contact on the left cup is determined by the \underline{\smash{active thermal sensor readings}}. The cup is assumed to not be in contact at the initial timestep. Contact is detected when the cup was not in contact on the previous timestep, and the active thermal difference reading is below -0.01 °C. We determine that contact is released when the sensor was in contact at the previous timestep, the active thermal rate is positive, and the temperature is below 53 °C. The condition on active thermal reading prevents false triggers caused by the on-off controller that maintains the temperature of the active thermal sensor.
    
    \item Contact on the right cup is determined by the \underline{\smash{force sensor readings}}. The cup is assumed to not be in contact at the initial timestep. Contact is detected when the cup was not in contact on the previous timestep, and the force signal exceeds a certain threshold. Contact is released when the sensor was in contact and the force falls below the same threshold. We set the threshold as 0.1V. 
\end{itemize}

Once all contact onsets and releases in a trial have been detected, the trial data is segmented to include only the periods of contact. All features shorter than 491 timesteps are padded. 

\subsection{Filtering dataset for model learning}
\label{appendix:filtering_dataset}

The CLAMP dataset contains many data points that are not directly usable for our task of material recognition. We exclude the following data points that are a part of our dataset, for model training:

\begin{itemize}
    \item Contact instances with objects of material \texttt{granite} and \texttt{dry wall}, the two smallest material classes. This helps to alleviate the class imbalance problem in material recognition. 
    \item Contact instances with non-zero force at the first timestep, which indicates that contact occurred before recording began.
    \item Contact instances where initial active temperature is less than 51 $\degree C$. Prior work has shown that material recognition is challenging under varying initial conditions \cite{bhattacharjee2015material}. 
    \item Contact instances where one of the temperature sensors shows erroneous readings. This occurs when the thermal sensors disconnect during deployment. We do not discard this data point from our dataset as data from other haptic modalities remain usable towards recognition.  
    \item Contact instances with maximum angular velocity below a threshold (chosen as 1$\degree/s$, indicating a very slow grasp.
    \item Contact instances with objects that have heterogeneous surfaces. 
\end{itemize}

\subsection{Data annotation}
\label{appendix:data_annotation}

We provide the following prompt to GPT-4o, to generate ground truth annotations for material. The system-level prompt is as follows:

\begin{lstlisting}[style = LLMQuery]
You are an oracle that is part of the Haptic Dataset project. Your role is to inform us what object it is, and all the materials that the object is made of. A reacher-grabber will grasp this object from the left and the right.
\end{lstlisting}

The prompt contains text along with more than 12 in-context examples containing text and images. The prompt is as follows:

\begin{lstlisting}[style = LLMQuery]
You will be provided with an image and a human-generated audio annotation converted to text as input. Respond in the following manner: 
1. Object: 
2. Materials: 
3. Heterogeneous Surfaces: 
Here are some rules : 
- Under the 'Object' section, specify the object that is being referred to using just the image and the prompt.
- Under the 'Materials' section, identify ONLY ONE material that the object is in contact with, directly or indirectly.Choose a material ONLY the following materials lists: [foam, aluminium, wood, steel, dry_wall, soft_plastic, hard_plastic, glass, paper, porcelain, granite, cardboard, rubber, vegetable_matter, fabric, brass].
- If the material specified by text is 'Unknown', use the image to recognize the materials.
- Under the 'heterogeneous Surfaces' section, specify if the object is made up of different materials on the two opposing sides
\end{lstlisting}

Each in-context example is provided in the following format (the horizontal line demarcates the two inputs sent as one user prompt and the assistant prompt containing the example annotation)

\begin{lstlisting}[style = LLMQuery]
Input audio transcription : It's a painted stainless steel cup.
Image : <encoded image> 
----------------------------------------------------------------------------
1. Object: cup 2. Materials: steel 3. heterogeneous Surfaces: No

\end{lstlisting}

\subsection{Dataset statistics}

\label{appendix:dataset_statistics}

We demonstrate the diversity in the CLAMP dataset along three axes: distribution of object materials, grasping forces, and grasping speeds. 
In Figure \ref{fig:material-distribution}, we show the spread in values of maximum force, maximum grasping speed, along with object material. Additionally, we plot the duration of contact instances (before padding) as a histogram.

For the material distribution, we only consider objects that do not have heterogeneous surfaces. Our dataset contains 5357 objects, out of which 680 ($\sim 13\%$) objects are objects with heterogeneous surfaces. We plot the distribution of material labels for contact instances with the remaining homogeneous objects. Figure \ref{fig:material-distribution} shows that the material labels contain significant class imbalance, with the largest class size (\texttt{hard plastic}) containing approximately 1000x as many samples as the smallest class (\texttt{dry wall}). 

\begin{figure*}[t]
\centering
\includegraphics[width=1\linewidth]{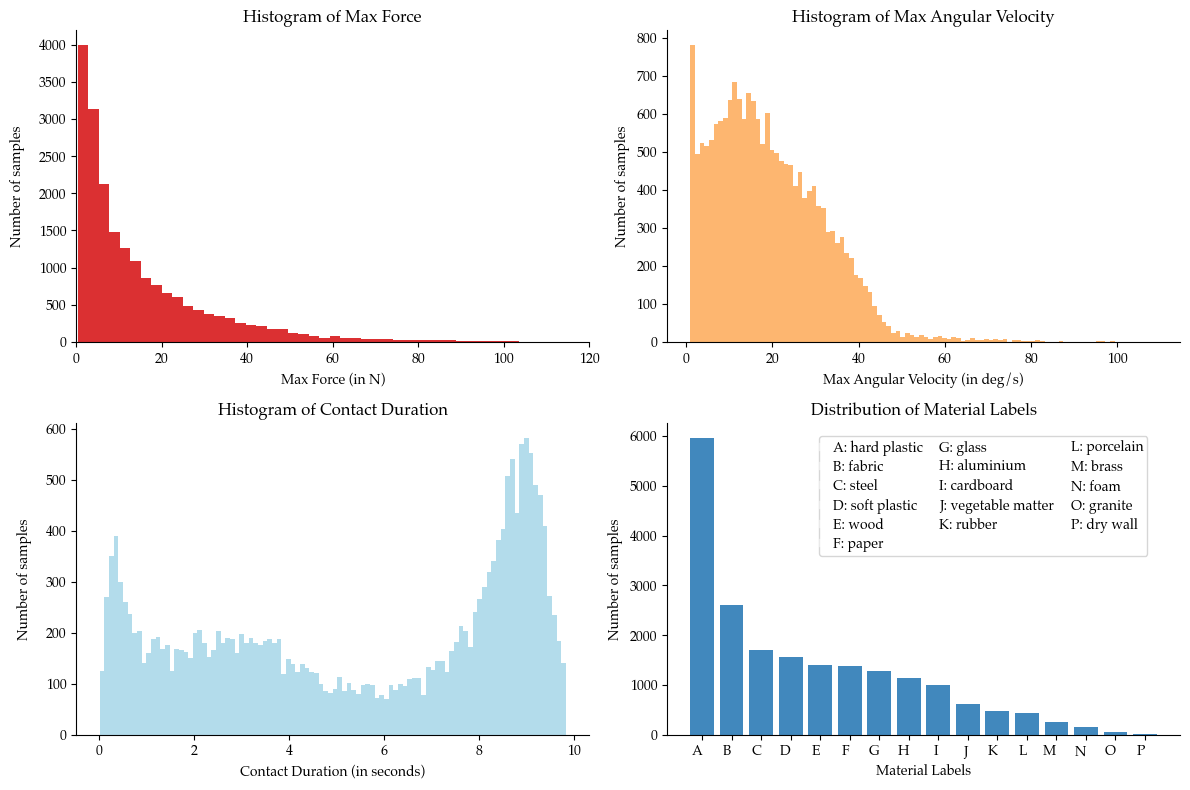}
\caption{Visualizing the diversity in the CLAMP Dataset along various axes}

\label{fig:material-distribution}
\end{figure*}

\section{The CLAMP model}

\subsection{Model training details}
\label{appendix:model_training}

\textbf{Hyperparameters and code details.} The hyperparameters for the three experiments: haptic encoder training, CLAMP model pretraining, and CLAMP model finetuning on robot data, are provided in Table \ref{tab:hyperparam}. The training code for the haptic encoder was written using the TSAI \cite{tsai} API, while that for the CLAMP model pretraining and finetuning was written using PyTorch API. 

\textbf{Haptic encoder details.} We use the InceptionTime architecture \cite{ismail2020inceptiontime}, choosing the number of filters as 256. Our model uses 7 1-dimensional convolutional kernels, the smallest and largest ones being of length 7 and 250 respectively, that are applied on each latent feature. These components make up one of the 6 Inception blocks in our model. The 6 blocks are followed by a 3-layer MLP that takes a latent embedding of dimension 2048 as input, and returns logits of dimension 14 as output. 

\textbf{Label weights.} We use label weights from the CLAMP pretraining dataset to train the haptic encoder and then the CLAMP model. Since the robot-collected finetuning set has a different material distribution, we recalculate weights as the inverse class frequencies of the finetuning data instead of reusing the pretraining weights, in our finetuning experiments. 

\textbf{Layers used for learning.} During pre-training on the CLAMP dataset, we freeze the haptic encoder. During CLAMP model finetuning, we continue to train the fusion MLP with the new label weights and unfreeze the haptic encoder to adapt it to haptic data from the robot embodiment. 

\textbf{Loss function used for training CLAMP model.} For the KL-divergence loss, we observe empirically that treating the vision distribution as the target and visuo-haptic as the student performs better than vice-versa.

\begin{table*}[tbp]
  \centering
  \scriptsize
  \setlength{\tabcolsep}{4pt} 
  \label{tab:modalities}
  \begin{tabular}{%
        @{}                                     
        >{\centering\arraybackslash}p{0.28\textwidth}
        | >{\centering\arraybackslash}p{0.28\textwidth}
        | >{\centering\arraybackslash}p{0.28\textwidth}
        @{}                      
    }
    \begin{tabular}[t]{@{}lcc@{}}
      \toprule
      \multicolumn{2}{c}{\bfseries Haptic encoder pre-training} \\
      \midrule
      Hyperparameter &  Value  \\
      \midrule
      Learning rate & 1e-5 \\
      Weight decay & 0 \\
      Filters in InceptionTime model & 256 \\
      Batch size & 64 \\
      Epochs & 100 \\
      \bottomrule
    \end{tabular}
    &
    \begin{tabular}[t]{@{}lcc@{}}
      \toprule
      \multicolumn{2}{c}{\bfseries CLAMP model pre-training} \\
      \midrule
      Hyperparameter &  Value  \\
      \midrule
      Learning rate & 1e-5 \\
      Weight decay & 0 \\
      Filters in InceptionTime model & 256 \\
      Batch size & 64 \\
      Epochs & 120 \\
      $\lambda_{KL}$ & 0.1 \\
      \bottomrule
    \end{tabular}
    &
    \begin{tabular}[t]{@{}lcc@{}}
      \toprule
      \multicolumn{2}{c}{\bfseries CLAMP model finetuning} \\
      \midrule
      Hyperparameter &  Value  \\
      \midrule
      Learning rate & 1e-5 \\
      Weight decay & 0 \\
      Filters in InceptionTime model & 256 \\
      Batch size & 64 \\
      Epochs & 30 \\
        $\lambda_{KL}$ & 0.1 \\
      \bottomrule
    \end{tabular}
  \end{tabular}
  \caption{Hyperparameters for learning experiments}
  \label{tab:hyperparam}
\end{table*}

\textbf{Seeds used for experiments.} We perform all our experiments on the seed, chosen out of 3 random seeds, on which the haptic encoder shows the worst performance. We use this seed, $18$, for all the experiments in this paper. 

\textbf{Experiments involving CLIP encoder.} We use the ViT-B32 version of the CLIP model for our experiments. We finetune the last two layers of the model and attach a classification head for the vision-only experiments. For the visuo-haptic model with CLIP as the vision head, we concatenate a latent embedding of dimension 512 from CLIP, with the embedding of dimension 2048 from the haptic encoder (from the layer before the haptic encoder classification head), and then pass it to the MLP that fuses these visual and haptic embeddings to generate a material prediction. 

\textbf{Unknown and uncertain predictions.} For the CLAMP model, we find that $p_1 = 0.45$ and $p_2 = 0.25$ strikes a good balance between increase in performance (across accuracy and nMCC) and number of rejected predictions. 

\subsection{Extracting Visual Features from GPT-4o}
\label{appendix:features_gpt}

We obtain log-probabilities from GPT-4o, renormalize them to redistribute probability mass across classes, and convert the results into logits for low-dimensional feature fusion.

To obtain a material prediction and log-probabilities, we employ a two-step prompting approach for better material classification. We initially feed in an image to identify the object in question. Then, we further prompt GPT with the predicted object and input image to generate a prediction for object material. In-context examples are provided for only the second step. For this step, we request the top 14 logits. We filter out logits with tokens that do not belong to the list of materials, and assign a probability of 0 to those material classes for which a log-probability was not generated. 

For the classes for which a log-probability was generated, we apply the softmax function and apply the fourth root. This step ensures a wider spread of probabilities across material classes. Finally, we normalize the transformed probabilities to generate logits with zero mean and unit standard deviation.

\section{Real robot experiments}

\begin{table}[!h]
  \centering
  \begin{tabular}{ c  c } 
    \toprule
    \textbf{Material} & \textbf{Trash/Recycle} \\
    \midrule
    Aluminium  & Recycle \\
    Brass  & Recycle \\
    Cardboard  & Recycle \\
    Fabric  & Trash \\
    Foam  & Trash \\
    Hard plastic & Trash \\
    Paper & Recycle \\
    Porcelain & Trash \\
    Rubber & Recycle \\
    Soft plastic & Trash \\
    Steel & Recycle \\
    Vegetable matter & Trash \\
    Wood & Trash \\
    \bottomrule
  \end{tabular}
  \vspace{3pt}
  \caption{Rules for trash sorting}
  
  \label{tab:sorting}
\end{table}

\subsection{Feature extraction for parallel jaw gripper embodiments}
\label{appendix:robot_finetuning}

Feature extraction for data from the Franka with CLAMP device embodiment is identical to that for data from CLAMP devices. Feature extraction for the other two robot embodiments differs only in the proprioception and impedance features. We use the linear acceleration of IMU-2 relative to IMU-1 along IMU-2’s Z-axis and process it to remove noise. First, we smooth the signal using a Savitzky-Golay filter. Then, we debias it using the mean of the last 50 timesteps, as the robot gripper is stationary near the end of data collection. Finally, we model and compensate for linear drift using readings from the first and last timestamps. This yields the proprioception feature for the parallel jaw gripper. 

For impedance, we use the ratio of the force difference to linear acceleration feature. A lower threshold is not needed as the robots grasp objects with predefined speeds. 

\subsection{Data collection for robot embodiments}
\label{appendix:embodiment}

For all real-robot experiments, we assume the grasp of each object. For the sorting and metallic object retrieval experiments, each object also has a pre-defined pose from where an image is captured. For objects that are unstable (e.g. cello tape), we hold them or weakly attach them to the table using adhesive. This is not dissimilar to human data collection as users report holding the object with one hand while grasping using the CLAMP device in the other hand. Finally, for each object, we capture an image using the CLAMP device camera, to create the dataset. 

The dataset for the ``Franka with CLAMP device" embodiment is 60 objects, while datasets for the parallel jaw embodiments is 30 objects each. Comparing finetuned model performance across embodiments, we hypothesize that ``Franka with CLAMP" underperforms ``Franka with Parallel Jaw" because it includes data from wide objects that the parallel jaw embodiment cannot grasp. Because the robot grasping policy was open-loop, this led to variable and noisy contact between the CLAMP device grippers and these wider objects (wider objects led to inconsistent and partial surface area contact between sensors and object surfaces). This was not the case during human data collection because humans used the LED signals to close the loop on their grasping policy such that the contact was complete and consistent. To test our hypothesis, we finetune the CLAMP model on 30\% data out of all data composed of only objects grasped by all embodiments. With 0.96 test accuracy, ``Franka with CLAMP" slightly outperforms ``Franka with Parallel Jaw" (0.95) and ``WidowX with Parallel Jaw" (0.90), which better reflects the effect of device and various robot embodiment differences.
\subsection{Sorting recyclable from non-recyclable items}
\label{appendix:robot_sorting}

All the objects are grasped in a fixed sequence, using pre-defined poses for image capture and grasp. The set of 10 objects is not changed across the 3 trials. We used an unknown prediction threshold $p_1 = 0.18$ and an uncertain prediction threshold of $p_2 = 0.04$. The rules governing sorting for this experiment, based on object material, are outlined in Table \ref{tab:sorting}.

\subsection{Retrieving objects from a cluttered bag}
\label{appendix:robot_bag}

For this experiment, we place three objects in a cloth bag: a T-shirt, a steel fork, and a bag of popcorn. The robot captures an image, reaches into the bag, grasps each object, and retrieves objects if the CLAMP model predicts it to be one of (\texttt{aluminium}, \texttt{brass}, or \texttt{steel}). We use a total of two object sequences across the 13 trials: fork $\rightarrow$ T-shirt $\rightarrow$ popcorn and T-shirt $\rightarrow$ fork $\rightarrow$ popcorn. We observe that grasping the bag first or second causes the cloth bag to crumple and thus prevent further grasps. 

\subsection{Separating ripe from overripe bananas}
\label{appendix:robot_banana}

We finetune the haptic encoder transferred to compliance classification, on haptic data from the Franka with Parallel Jaw embodiment. We use the same hyperparameters as those used for finetuning in Table \ref{tab:hyperparam}. For this experiment, we classified bananas as ripe if the compliance prediction was \texttt{hard}, and overripe if the prediction was \texttt{soft}.

\end{document}